\documentclass[conference]{IEEEtran}
\IEEEoverridecommandlockouts
\usepackage{cite}
\usepackage{amsmath,amssymb,amsfonts,bm}
\usepackage{algorithmic}
\usepackage{graphicx}
\usepackage{textcomp}
\usepackage{xcolor}
\usepackage{multirow}
\usepackage{hyperref}

\usepackage{epstopdf}

\def\BibTeX{{\rm B\kern-.05em{\sc i\kern-.025em b}\kern-.08em
    T\kern-.1667em\lower.7ex\hbox{E}\kern-.125emX}}
\begin{document}

\title{Tiny-YOLO object detection supplemented with geometrical data}

\author{\IEEEauthorblockN{1\textsuperscript{st} Ivan Khokhlov}
\IEEEauthorblockA{\textit{Wave Processes and Control} \\
\textit{Systems Lab, MIPT}\\
Moscow, Russia \\
khohklov.iyu@gmail.com}
\and
\IEEEauthorblockN{2\textsuperscript{nd} Egor Davydenko}
\IEEEauthorblockA{\textit{Wave Processes and Control} \\
\textit{Systems Lab, MIPT}\\
Moscow, Russia\\
egordv@gmail.com}
\and
\IEEEauthorblockN{3\textsuperscript{rd} Ilya Osokin}
\IEEEauthorblockA{\textit{Wave Processes and Control} \\
\textit{Systems Lab, MIPT}\\
Moscow, Russia\\
kefir8888@gmail.com}
\and
\IEEEauthorblockN{4\textsuperscript{th} Ilya Ryakin}
\IEEEauthorblockA{\textit{Wave Processes and Control} \\
\textit{Systems Lab, MIPT}\\
Moscow, Russia\\
ryakin.is@phystech.edu}
\and
\IEEEauthorblockN{5\textsuperscript{th} Azer Babaev}
\IEEEauthorblockA{\textit{Wave Processes and Control} \\
\textit{Systems Lab, MIPT}\\
Moscow, Russia\\
7684067@mail.ru}
\and
\IEEEauthorblockN{6\textsuperscript{th} Vladimir Litvinenko}
\IEEEauthorblockA{\textit{Wave Processes and Control} \\
\textit{Systems Lab, MIPT}\\
Moscow, Russia\\
litvinenko.vv@phystech.edu}
\and 
\IEEEauthorblockN{7\textsuperscript{th} Roman Gorbachev}
\IEEEauthorblockA{\textit{Wave Processes and Control} \\
\textit{Systems Lab, MIPT}\\
Moscow, Russia\\
}
}

\maketitle

\begin{abstract}
%This paper presents our research results in object detection problem using geometrical data: all objects are situated on the same plane. We obtain and mark a new dataset with images and camera position relative to the plane. We suppose that this additional information will significantly improve the detection quality.

We propose a method of improving detection precision (mAP) with the help of the prior knowledge about the scene geometry: we assume the scene to be a plane with objects placed on it. We focus our attention on autonomous robots, so given the robot's dimensions and the inclination angles of the camera, it is possible to predict the spatial scale for each pixel of the input frame. With slightly modified \textit{YOLOv3-tiny} we demonstrate that the detection supplemented by the scale channel, further referred as \textbf{S}, outperforms standard RGB-based detection with small computational overhead.

\end{abstract}


\section{Introduction}

\subsection{Research Area Overview}

Our interest is mainly focused on autonomous bipedal robots. We validate our algorithms and ideas by taking part in world competition - RoboCup. The competition is a soccer game between two teams of autonomously playing robots. Each team consists of four robots maximum. Robots are supposed be similar to the humans in sensors, body structure, proportions and center of mass position. They can communicate via Wi-Fi network with each other and the referee. The playing area looks like FIFA football field, but is much smaller - $6 \times 9$ meters.

The robots' ability to play increases significantly each year. Our robots can localize and move pretty good and stable. The next huge step to improve the level of the game is to implement opponent robots detection. Also, it is important to increase quality and distance of ball detection. These two points push us to investigate methods of object detection in our specific conditions.

\subsection{Problem Statement}

We have mono camera FLIR Blackfly GigE\footnote{\url{https://www.flir.com/products/blackfly-gige}} with wide-angle lens and Intel NUC Core i5\footnote{\url{https://www.intel.ru/content/www/ru/ru/products/boards-kits/nuc/kits/nuc7i5bnk.html}} as computational platform\cite{tdp}. We need to detect opponent robots and ball on the playing field. Obstacle detection is an important part of navigation stack of humanoid soccer robot. It allows robot to behave much more like human player and not to hit other robots during the game. For this goal we can call “obstacle” any object on the field that cannot be bypassed by the robot. This usually includes opponent robots, ball and human referee. This can be reformulated to “an obstacle is any part of the playing field which is not green, and it’s not a field marking line”. This formalization of an obstacle can be turned into a simple classic computer vision pipeline that is capable of running in real time on the robot for obstacle detection as described below. The second approach to this problem is object detection using neural networks. This will give us more information about the current field state. We can obtain information about number of opponent robots and their coordinates. The application of this approach is more promising because it will give us new data to improve strategy. It is important that we are strongly limited by $CPU$ computational abilities in neural network-related tasks. 

\subsection{Related work}

In the \cite{xu2019geometry} paper authors work in similar setting, but they obtain the geometrical data from the video stream. In this work a static camera is used, which is not the case in the autonomous robots competitions.

\cite{cai2019improving}
paper presents an approach basing on rectifying the image before detection. It allows the neural network to not to learn features for the distorted image. For the problem of detecting distant objects captured by a camera of the robot such a rectification is not possible due to the heavy distortions arising.

The approach in \cite{gupta2014learning} relies on the joint analysis of RGB and Depth images. Contours and region proposals are used to build depth encoding, and after the feature extraction fused with the RGB features. This approach cannot be adopted to the considered problem since the real Depth data is not present.

\section{Distance transform-based approach}
\begin{figure}
\begin{minipage}[h]{0.45\linewidth}
\center{\includegraphics[width=1\linewidth]{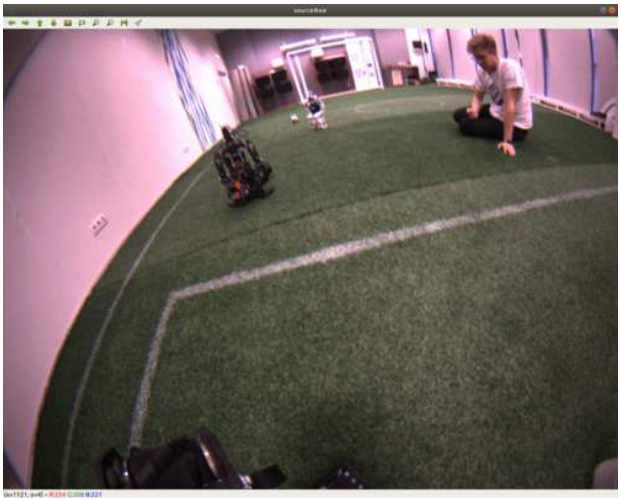}}  \\
\end{minipage}
\hfill
\begin{minipage}[h]{0.45\linewidth}
\center{\includegraphics[width=1\linewidth]{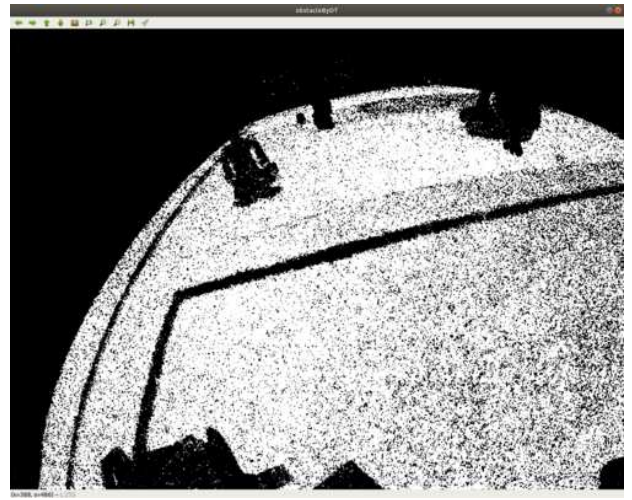}} \\
\end{minipage}
\caption{Input frame from robot’s camera and the output of “green filter”.}
\label{fig:one}
\end{figure}

Since the obstacle is formalized as part of the playing field that is not green,
it is possible to directly discriminate pixels of an input image to “green” and “not green” using so called “green filter” classifier that consists of $RGB$ to $HSV$ domain color transformation and thresholding. The resulting image of this filter is shown in the Fig. \ref{fig:one}.

The sample input image contains different types of not green objects on the field: two opponent robots (black and white), human referee, and field marking lines. All of these objects are correctly classified as “not green” by the green filter. The resultant image is further processed by morphological operations (erosion and dilation)\cite{hipr} to filter out noisy pixels that appear as fake green on non-green objects.

\begin{figure}
\begin{minipage}[h]{0.45\linewidth}
\center{\includegraphics[width=1\linewidth]{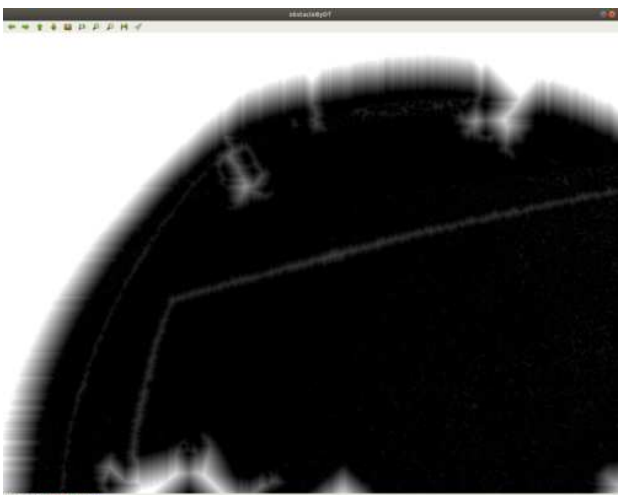}}  \\
\end{minipage}
\hfill
\begin{minipage}[h]{0.45\linewidth}
\center{\includegraphics[width=1\linewidth]{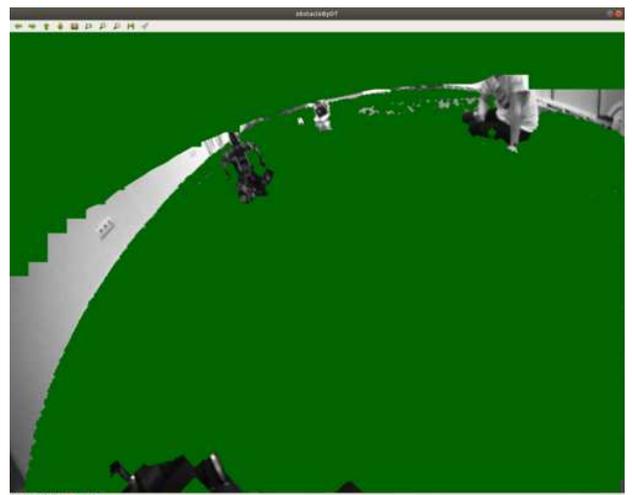}}  \\
\end{minipage}
\caption{Distance transform output (left) and its thresholding result using knowledge of assumed field marking width at any source image pixel (right). Green colored area is walkable or too far.}
\label{fig:2}
\end{figure}

The next step is to distinguish marking lines among the other objects. For filtering the field markings we use the prior knowledge of the width of a marking line that is constant in any part of the field. The assumed line marking width in pixels can be calculated for all input image pixels using the data about camera orientation relatively to the field planar surface available in real-time from robot’s internal model subsystem. To filter objects that are thinner than the field marking line we use the distance transform approach\cite{hipr}. The result of the distance transform applied to the black and white image is a single-channel image in which the value of the pixel is equal to the distance to the closest non-zero pixel in source image. The result of the distance transform of a sample green filter output can be seen in Fig. \ref{fig:2}(left). By thresholding this image with the help of pixel-wise knowledge of field line marking width we come to the result shown in Fig. \ref{fig:2}(right). The available for walking areas are shown in green color.

\begin{figure}
    \begin{minipage}[h]{\linewidth}
    \center{\includegraphics[width=1\textwidth]{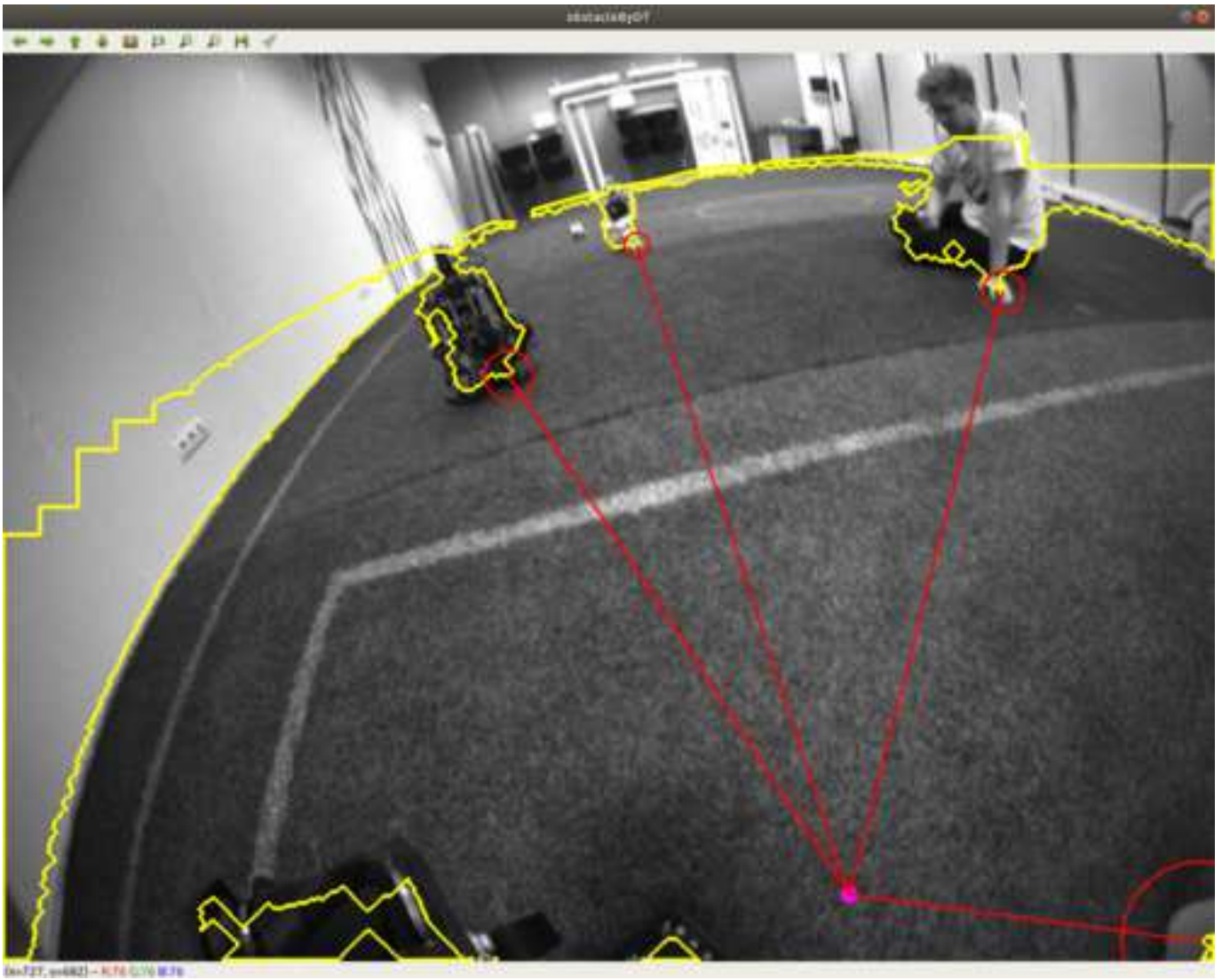}}
    \caption{Distance transform clusters boundary marked by yellow. Boundary pixels classified as a part of an obstacle marked by red lines pointing from the robot’s next step position on the field marked by magenta dot.}
    \end{minipage}
    \label{fig:3}
    \hfill
    % \begin{minipage}[h]{0.3\linewidth}
    % \center{\includegraphics[width=1\textwidth]{imgs/fig6.eps}}
    % \caption{Good example of the inference of the net trained in the regular way on the test set.}
    % \label{fig:4}
    % \end{minipage}
    % \hfill
    % \begin{minipage}[h]{0.3\linewidth}
    % \center{\includegraphics[width=1\textwidth]{imgs/fig7.eps}}
    % \caption{Scale-supplemented net confuses two types of balls, neglecting the color data prior to the scale data.}
    % \label{fig:5}
    % \end{minipage}
\end{figure}

The last part of the filtering is the boundary detection of the non-zero clusters found on the thresholded distance transform result. This boundary is shown by yellow lines in Fig. \ref{fig:3}. Each pixel of this boundary is then checked by the obstacle filter, and if it is close enough to the robot, it is treated as an obstacle and then processed in robot’s navigation stack.
	
\section{Deep Learning Approach}

\subsection{Dataset, model, augmentation}
For this approach we have used $YOLOv3-tiny$ DNN\cite{yolov3} as a baseline for the object detection. The choice was dictated by the conditions the net is deployed in and the properties of the architecture itself, meaning only $CPU$ (no graphic accelerator on board), relatively small available dataset, small number of classes and required real-time or almost real-time inference.

We have collected and marked a real-world dataset of $1705$ frames from the camera of the robot (see section Problem Formulation for camera specification), considering the following classes: ball (standard Robocup ball), goal post, rhoban (the type of robots that we use).

We have used the following augmentation methods: vertical and horizontal translation (± 10\%), rotation (± 5 degrees), vertical and horizontal shear (±2 degrees), scale (±10\%), horizontal reflection, HSV Saturation and Value change (±50\%). The augmentation was applied during training.

\subsection{Geometrical data incorporation}
We have attempted to exploit the specific properties of the problem, particularly the knowledge about the scale of every pixel in the scene. Knowing the inclination angles of the camera, the spatial dimensions of the robot and the calibration matrix of the camera it is possible to obtain the scale on the field for every pixel. This scale data that is equivalent to the depth map can be present in the form of a single channel image of the same dimensions as the original.

Since the scene is assumed to be a plane, it is possible to interpolate \textbf{S} between pixels with the stride of $32$ without loss of quality, which makes the computational overhead for calculating \textbf{S} very small. Experiments show that inferring $4$-channel net on $1088 \times 1440$ images is approximately $1.232$ times more time-consuming than $3$-channel.

%In order to keep using pretrained three-channel net we have replaced green channel in each image with the scale data. Green channel was chosen to be replaced since red and blue channels are the most distant from each other in terms of the spectrum and they are expected to be the least correlated, thus the most informative among all the pairs of the channels.

The intuition behind this approach is that $YOLO$ can probably generalize the connection between the size of the object and the class of the object and improve the quality of the predictions.

\subsection{Current results, discussion}

\begin{table}[h!]
\caption{Comparison with different IoU between YOLOv3-tiny RGB and RGB-\textbf{S}  for \textbf{mAP} metric}
%\caption{}
%\label{tab:comparison}
\begin{center}

\begin{tabular}{ |c|c|c| } 
 \hline 
 & \multicolumn{2}{c|}{\textbf{YOLOv3-tiny}} \\
 \hline 
 & \textbf{RGB} & \textbf{RGB-S} \\ 
 \hline
 %\bf{F1} & 0.84 & 0.872 \\ 
 %\hline
 \textbf{mAP@0.5} & 0.84 & 0.872 \\ 
 \hline
 \textbf{mAP@0.6} & 0.819 & 0.867 \\ 
 \hline
 \textbf{mAP@0.7} & 0.766 & 0.818 \\ 
 \hline
 \textbf{mAP@0.8} & 0.638 & 0.734 \\ 
 \hline
 \textbf{mAP@0.9} & 0.503 & 0.599 \\ 
 \hline
\end{tabular}
\end{center}
\label{tab:comparison}
\end{table}

\subsection{Future work}

There are many approaches and hypothesis to be tried out in the chosen direction of adding geometrical data directly to the input data, among them we can highlight the following:

\begin{itemize}
    \item Try regular (not tiny) YOLO, Spiking YOLO \cite{spiking-yolo} and another architectures.
    \item Try less straightforward adding of the scale data, for instance add few independent convolutional layers before fusing it with the color data.
\end{itemize}

\section{Conclusion}

The robot's behaviour strongly relies on the precise object detection, since the distances to the objects are calculated out of their coordinates in the frame. Table \ref{tab:comparison} shows that the proposed method significantly outperforms baseline with high demands to the detection precision, see the \textbf{mAP@0.9} line.

The quality improvement is achieved with relatively small computational overhead, which is critically important for the autonomous robots due to the limited resources. The method's application does not require any extra data apart from \textbf{S}, calculated from the robot's parameters and measurements that were already present. The incorporation of the geometrical data was made on the setup with a single RGB camera.

The proposed approach is not limited to the autonomous robotics applications, it can be used to improve detection accuracy when the scene geometry is known.

\bibliographystyle{./bibliography/IEEEtran}
\bibliography{./bibliography/IEEEabrv,./bibliography/IEEEexample, ./bibliography/references}

\newpage
\clearpage
\onecolumn

\begin{figure}
\begin{minipage}[h]{0.45\linewidth}
\center{\includegraphics[width=1\linewidth]{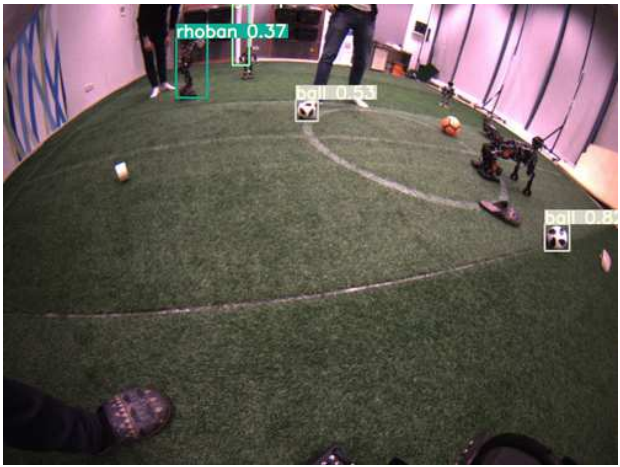}}  \\
\end{minipage}
\hfill
\begin{minipage}[h]{0.45\linewidth}
\center{\includegraphics[width=1\linewidth]{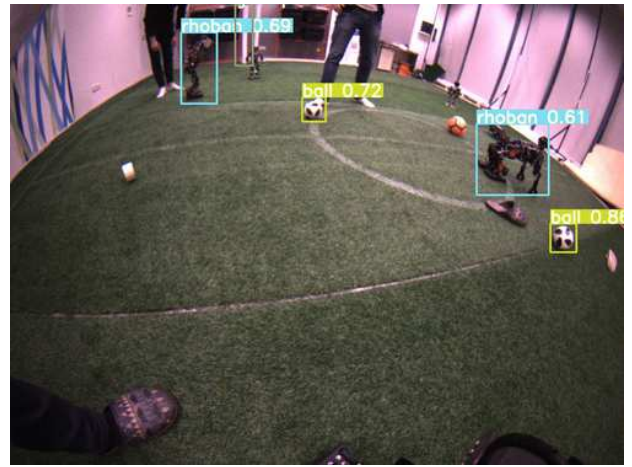}} \\
\end{minipage}
\caption{Left image - RGB, right - RGB-\textbf{S}, note that rhoban in the right side of the field is missing in the first of them. Also \textbf{S} channel improves the bbox accuracy, see left rhoban occurrence.}
\label{fig:yolo1}
\end{figure}

\vfill

\begin{figure}
\begin{minipage}[h]{0.45\linewidth}
\center{\includegraphics[width=1\linewidth]{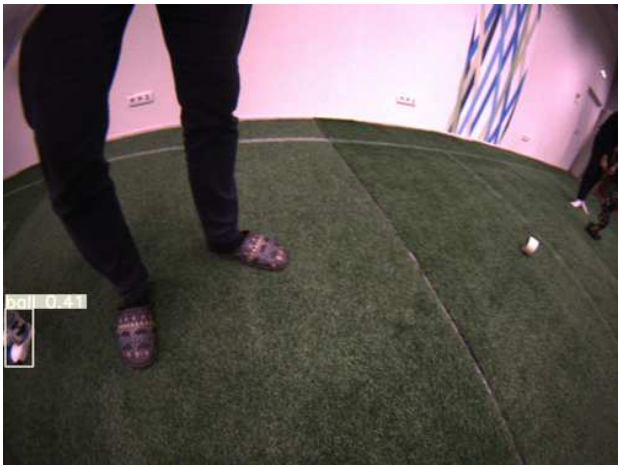}}  \\
\end{minipage}
\hfill
\begin{minipage}[h]{0.45\linewidth}
\center{\includegraphics[width=1\linewidth]{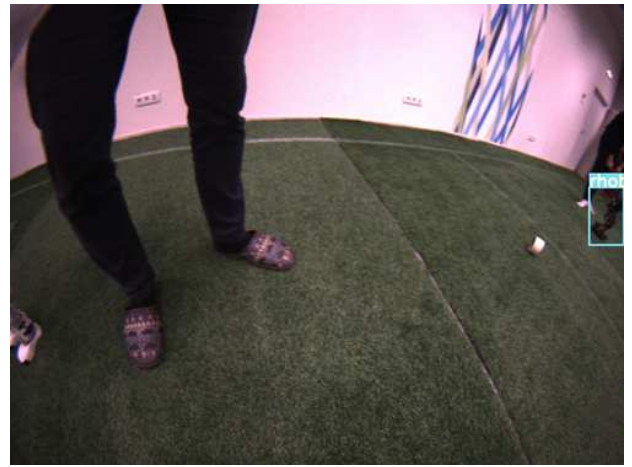}} \\
\end{minipage}
\caption{Left image - RGB, right - RGB-\textbf{S}. These images demonstrate improvement not only in the detection precision, but also in classification: in the left the foot of Nao v.6 robot is confused with the ball, while the rhoban in the right side of the frame is missing.}
\label{fig:yolo2}
\end{figure}

\vfill

\vfill

\vfill

\end{document}